%% file: PaperForReview.tex
\documentclass[10pt,twocolumn,letterpaper]{article}

\usepackage[pagenumbers]{cvpr} % To force page numbers, e.g. for an arXiv version
\makeatletter
\@namedef{ver@everyshi.sty}{}
\makeatother
\PassOptionsToPackage{dvipsnames}{xcolor}
	\RequirePackage{xcolor} % [dvipsnames]
\usepackage{graphicx}
\usepackage{booktabs}
\usepackage{amsthm}
\usepackage{amsmath}
\usepackage{amsfonts}
\usepackage{amssymb}
\usepackage{bm}
\usepackage{mathtools}
\usepackage{dsfont}
\usepackage{empheq}
\usepackage{stmaryrd}
\usepackage{nicefrac}

\usepackage{adjustbox}
\usepackage{pifont}
\usepackage{multirow}

\usepackage{mathrsfs}

\usepackage{graphbox}
\usepackage{tikz}
\usepackage{pgfplots}
\pgfplotsset{compat=newest}
\usepackage{caption}
\usepackage{subcaption}
\usepackage{standalone}
\usepackage{flushend}

\usepackage{tabularx}

\usepackage{adjustbox}

\usepackage{rotating}
\usepackage{multirow}
\usepackage{array}
\usepackage{booktabs}

\usepackage{lipsum}
\usepackage{colortbl}
\usepackage{footmisc}
\usepackage{setspace}

\usepackage[pagebackref,breaklinks,colorlinks,citecolor=blue]{hyperref}

\def\cvprPaperID{15} %Enter the CVPR Paper ID here
\def\confName{CVPR}
\def\confYear{2022}

\input{macro}

\begin{document}

\title{Multi-Head Distillation for Continual Unsupervised Domain Adaptation \\in Semantic Segmentation}

\author{Antoine Saporta\textsuperscript{1,2}
\quad
Arthur Douillard\textsuperscript{1,3}
\quad
Tuan-Hung Vu\textsuperscript{2}
\quad
Patrick Pérez\textsuperscript{2}
\quad
Matthieu Cord\textsuperscript{1,2}
\\
\textsuperscript{1}Sorbonne Université
\qquad
\textsuperscript{2}valeo.ai
\qquad
\textsuperscript{3}Heuritech
\\
{\tt\small antoine.saporta@isir.upmc.fr, arthur.douillard@heuritech.com,}\\ 
{\tt\small \{tuan-hung.vu, patrick.perez, matthieu.cord\}@valeo.com}
}
\maketitle

%%%%%%%%% ABSTRACT
\begin{abstract}
   Unsupervised Domain Adaptation (UDA) is a transfer learning task which aims at training on an unlabeled target domain by leveraging a labeled source domain. Beyond the traditional scope of UDA with a single source domain and a single target domain, real-world perception systems face a variety of scenarios to handle, from varying lighting conditions to many cities around the world. In this context, UDAs with several domains increase the challenges with the addition of distribution shifts within the different target domains. 
   This work focuses on a novel framework for learning UDA, continuous UDA, in which models operate on multiple target domains discovered sequentially, without access to previous target domains. We propose MuHDi, for Multi-Head Distillation, a method that solves the catastrophic forgetting problem, inherent in continual learning tasks. MuHDi performs distillation at multiple levels from the previous model as well as an auxiliary target-specialist segmentation head. We report both extensive ablation and experiments on challenging multi-target UDA semantic segmentation benchmarks to validate the proposed learning scheme and architecture.
\end{abstract}

%%%%%%%%% BODY TEXT
\section{Introduction}
\label{sec:intro}
\input{figures/bigpicture.tex}
Autonomous vehicles have recently returned to the forefront with the dazzling progress of AI and deep learning as well as the emergence of many players around new forms of mobility. %The recent progress in AI brought by deep learning largely explains the spectacular resurgence of driverless cars. 
Thanks to the new generation of convolutional neural networks, cameras may be embedded to understand, in real-time, crucial aspects of the environment: nature and position of vehicles, pedestrians and stationary objects; position and meaning of lane markings, signs, traffic lights; drivable area; etc. 

Today, these deep neural networks are trained in a fully-supervised fashion, requiring massive amounts of labeled data. While powerful, this form of training raises major issues. Collecting large and diverse enough labeled datasets for supervised learning is a complex and expensive undertaking. Moreover, such datasets remain limited, considering the diversity, complexity, and unpredictability of environments a vehicle may encounter. 
%Designing perception systems that may identify novel environments and continuously learn and adapt to changes, are some of the many challenges facing autonomous driving. 
%In this context, a prime tool is \textbf{Unsupervised Domain Adaptation (UDA)}. 
A prime tool to tackle these issues is \textbf{Unsupervised Domain Adaptation (UDA)}. 
Its goal is to adapt to a set of unlabeled data, the \emph{target domain}, sharing structures with another labeled dataset, the \emph{source domain}, allowing supervised training though having some statistical distribution differences. Practically, one could, for example, use UDA to train a model on unlabeled nighttime data as the target domain by taking advantage of labeled daytime data. Similarly, a promising practice is to use UDA to leverage synthetic data as the source domain, on which annotation is automatic and cheap, to train models on real-world data as the target domain. 

%In the past few years, UDA has been extensively studied by the community, especially for the semantic segmentation task. 
While the standard UDA scenario, with one source domain and one target domain, is useful for autonomous driving applications, especially in the synthetic-to-real context, it is still very limited when considering practical use-cases. Indeed, autonomous vehicles may encounter a large variety of urban scene scenarios in the wild, such as varying weather conditions, lighting conditions, or different cities, each of those representing a specific domain. 
%While standard UDA allows us to train a model on a particular unlabeled target domain, it may not generalize to this wide variety of scenarios. Furthermore, autonomous systems should be able to continuously improve and adapt to new target domains
While standard UDA allows us to train a model on a particular unlabeled target domain, it may not be able to continuously improve and adapt to new target domains. 
%, for instance when deploying autonomous vehicles to a new country
%. However, keeping access to previous real-world data is often unfeasible for various reasons such as privacy and perception systems must avoid forgetting about the domains it has previously seen when training on a new target domain.
Moreover, keeping access to previous real-world data is often unfeasible for various reasons, such as privacy, and a perception system must avoid forgetting about the domains it has previously seen when training on a new target domain.

In this work, we study a continual learning-inspired setting of UDA, which we call \textbf{Continual UDA}. The goal is to sequentially train a model on new target domains, one at a time, while maintaining its performance on the previous target domains without forgetting. \autoref{fig:bigpicture} illustrates the differences between this novel setting and traditional and multi-target UDA. While multi-target UDA already extends traditional UDA to learn multiple unlabeled target domains at once, continual UDA is a more challenging setting in which the model must sequentially learn the new target domains, one domain at a time, and is evaluated on all the target domains, as in multi-target. The contributions of our work are threefold:
\begin{itemize}
    \item We propose a new continual learning task for UDA with experimental benchmarks based on popular semantic segmentation datasets in the UDA community;
    \item To solve this new problem, we develop MuHDi, for Multi-Head Distillation, which explictly solves the catastrophic forgetting problem by performing probability distribution distillation at multiple levels from the previous models as well as an auxiliary target-specialist segmentation head;
    \item We demonstrate the performance of the proposed approach compared to natural baselines as well as multi-target approaches, in which the target domains are all available simultaneously.
\end{itemize}

\section{Related Works}
\paragraph{Traditional Unsupervised Domain Adaptation.}
UDA has been an active research topic for a few years. Its objective is to learn on an unlabeled target domain $\mfrk{D}_t$ by leveraging a labeled source domain $\mfrk{D}_s$. 
The main challenge %of this task 
is to overcome the distribution shifts between the two domains. UDA generally relies either on aligning in some way the source and target features of the model to make them indistinguishable or on finding a transformation from source to target domain and using this transformation to train the model entirely in the target domain using the transformed source data. Different paradigms and strategies can be found in the literature to tackle the adaptation problem. 

Direct distribution alignment approaches \cite{gretton2012mmd,tzeng2014ddcn,long2015dan,long2017joint,sun2016deep} directly minimize some distance or a measure of discrepancy between the two domains. 
Image-level approaches build transformations at the image level, either by directly translating the source image into the target domain before the neural network model dedicated to solving the task \cite{bousmalis2017pixeldan,Hoffman_cycada2017,wu2018dcan,yang2020fda} or by finding representations that allow reconstructing of images in either domain, independently of their domain of origin \cite{bousmalis2016dsn,chang-cvpr2019,murez2017translation}. 
Finally, the last class of UDA approaches is based on adversarial training to align the distributions of some representations of source and target domains. Usually, the representations selected for adversarial alignment are either deep features extracted after the deep neural network backbone of the model right before the last layers dedicated to the task \cite{ganin2015reversegrad,tzeng2017adda,hoffman-arxiv2016} or directly close-to-prediction representations at the output of the model \cite{tsai-cvpr2018,vu-cvpr2019}, which is the most popular and effective strategy today for semantic segmentation.
 
\paragraph{Multi-Domain Learning.}
Multi-domain learning is tackled in the literature under a variety of settings. Domain generalization \cite{li2018domain,matsuura2020domain,zhao2020domain,pandey2021generalization} is a challenging task close to the DA problem. Its objective is to learn a model from one or several training (source) domains that will be able to generalize to unseen testing (target) domains. While very close to a multi-domain UDA, domain generalization differs from it by the total absence of target data during training, leading to drastically different approaches to tackle these tasks.

Multi-source UDA \cite{zhao2019multisource,he2021multi,nguyen2021stem} aims at training using an arbitrary number of source domains to learn better, more generalizable features for a single target domain. While one could train a model with a standard UDA approach on the combination of all the considered source domains, it proves to be inefficient due to the discrepancy between the domains. Multi-source approaches effectively account for these distribution shifts and leverage them to produce more generalizable features.

In Open-Compound Domain Adaptation (OCDA) \cite{Liu_2020_CVPR,park2020discover}, the target domain may be considered as a combination of multiple homogeneous target domains -- for instance, similar weather conditions such as `sunny', `foggy', etc. -- where the domain labels are not known during training. Moreover, previously unseen target domains may be encountered during inference. 

\paragraph{Multi-Target Unsupervised Domain Adaptation.}
Multi-target UDA is still a fairly recent setting in the literature and mostly tackles classification tasks. While multi-target UDA is close to the OCDA setting in that it considers multiple target domains instead of a single one, multi-target UDA differs from this last setting in that it assumes that the domain of origin is known during training and that no new domains are faced at test time. Thus, for instance, multi-target UDA would be better suited to DA to multiple cities than OCDA since the origin of the training data should be easily accessible.
Two main scenarios emerge in the works on this task. In the first one, even though the target is considered composed of multiple domains with gaps and misalignments, the domain labels are unknown during training and test. 
\cite{peng2019domain} proposes an architecture that extracts domain-invariant features by performing source-target domain disentanglement. 
Moreover, it also removes class-irrelevant features by adding a class disentanglement loss. 
In a similar setting, 
\cite{chen2019blending} presents an adversarial meta-adaptation network that both aligns source with mixed-target features and uses an unsupervised meta-learner to partition the target inputs into clusters that are adversarially aligned. 
In the second scenario, the target identities are labeled on the training samples but remain unknown during inference. 
To handle it, \cite{yu2018multi} learns a common parameter dictionary from the different target domains and extracts the target model parameters by sparse representation; 
\cite{gholami2020unsupervised} adopts a disentanglement strategy by separately capturing both domain-specific private features and feature representations by learning a domain classifier and a class label predictor, and trains a shared decoder to reconstruct the input sample from those disentangled representations.

Tackling multi-target UDA in semantic segmentation has been proposed in two recent works. \cite{isobe2021multi} trains multiple semantic segmentation models, each one expert on a specific domain. These domain-specific expert models collaborate by being trained on images from the other domains stylized in the domain of expertise while making sure that the predicted maps are coherent between the experts for a same original image. Finally, the knowledge of all these experts is transferred to another model, which serves as domain-generic student. \cite{saporta2021mtaf} first proposes an approach combining two adversarial pipelines: the first one aims at discriminating the source domain to each individual target domain; the second one aims at discriminating each individual target domain to the other target domains. \cite{saporta2021mtaf} then proposes a second strategy adopting a multi-teacher/single-student distillation approach to learn a segmentation model which is agnostic to the target domains. 

\paragraph{Continual Learning.}
The task of continual learning aims at learning a constantly changing distribution. A naive mitigation is to re-train the model from scratch on the updated dataset. However, it assumes that previous data is kept, which is often unfeasible for multiple reasons, including privacy. Thus, a continual model has to learn solely on the new data while remembering the previous data. As a result, the model faces the challenge of ``catastrophic forgetting'' \cite{robins1995catastrophicforgetting,thrun1998lifelonglearning,french1999catastrophicforgetting} where the performance on previous samples drops. This problem can be mitigated by different approaches: rehearsal of a limited amount of previous data \cite{rebuffi2017icarl,hayes2020remind} can reduce forgetting, but is memory-costly for high-resolution images required for semantic segmentation \cite{douillard2021objectrehearsal}. A second approach is to constrain the new model to be ``similar'' to the previous model. This similarity can be defined on the weights \cite{kirkpatrick2017ewc}, the gradients \cite{lopezpaz2017gem}, or even the probabilities \cite{li2018lwf} and the intermediary features \cite{douillard2020podnet}. More recently, continual models were adapted for semantic segmentation \cite{michieli2019ilt,cermelli2020modelingthebackground,douillard2021plop} with success, but they restricted themselves to supervised tasks on a single dataset and not unsupervised adaptation across multiple domains. 

Closer to UDA, \cite{volpi2021continual} proposes a continual learning-inspired setting to the domain adaptation problem: a model is learned on multiple labeled domains seen sequentially which are all considered for evaluation. This setting, being  fully supervised, substantially differs from the novel setting proposed in our work, which is unsupervised for all the target domains and only leverages a labeled source domain. \cite{rostami2021lifelong} studies a UDA setting closer to ours. The model is sequentially trained on unlabeled target domains but requires storage of representative samples of the previous domains which are used for rehearsal. Furthermore, \cite{volpi2021continual,rostami2021lifelong} consider classification tasks while the focus of our work is semantic segmentation. 

\section{Continual UDA Setting}
We define in this section the continual unsupervised domain adaptation problem, inspired by continual learning tasks, and we propose extensions of traditional unsupervised domain adaptation methods as baselines for this new task. This problem is illustrated in \autoref{fig:bigpicture}.
\paragraph{Problem Setting.}
Let us consider $T\geqslant 2$ distinct target domains $\mfrk{D}_{t}$, $t\in [T] = \left\lbrace i\in \mbb{N} \mid 1 \leqslant i \leqslant T \right\rbrace$, to be jointly handled by the model. They are represented by unlabeled training sets $\domain_{t}$, $t\in [T]$. As in traditional UDA settings, we assume that the labeled training examples $(\vx,\vy)\in \domain_s\times\mcl{Y}_s$ stem from a single source domain, a specific synthetic environment for instance.

This setting differs from the multi-target UDA setting: we assume here that the different target domains can only be accessed sequentially and one at a time during training. More precisely, during step $t\in [T]$ of the training, the model only has access to a single target dataset, $\domain_{t}$, and cannot access ever again the target datasets $\domain_{k},\: k\in[t-1]$, it has previously learned. Nevertheless, we consider that the source domain dataset $(\domain_s,\mcl{Y}_s)$, which is often synthetic, is always accessible during training.
We denote $F_{(1:t)}$ the continual model that one obtains after sequentially training from target domains $1$ to $t$.

%However, t
The objective of continual DA is the same as in a multi-target setting:  train a single semantic segmentation model, $F_{(1:T)}$, that achieves equally good results on all target-domain test sets. 
Also, while the target domain of origin is known for all unlabeled training examples, we assume, as in multi-target approaches of \cite{gholami2020unsupervised,yu2018multi,isobe2021multi,saporta2021mtaf}, that this information is not accessible at test time.

The continual DA setting brings a new challenge compared to the multi-target one: the model must not forget the previous target domains it has learned before (\textit{rigidity}) while still being able to adapt to new domains (\textit{plasticity}).

\paragraph{Revisiting Adversarial UDA.}
We adapt the training procedure of state-of-the-art UDA approaches like \cite{tsai-cvpr2018} or \cite{vu-cvpr2019} to this new setting.
In continual UDA, the model always has access during training to a single source domain and a single target domain. Traditional UDA approaches can easily be adapted to this setting: at each iteration $t\in[T]$, the model is trained on the source dataset $\domain_s$ and the current target dataset $\domain_{t}$, initialized from the model trained at the previous iteration. This way, the model is trained in a UDA fashion sequentially on all the target domains. However, this ``continual baseline'' does not directly tackle catastrophic forgetting of old domains. 
The following section describes our strategy to explicitly prevent catastrophic forgetting.  

\section{Multi-Head Distillation Framework}
This section presents our approach to continual UDA for semantic segmentation: MuHDi, for Multi-Head Distillation.
%\autoref{fig:md_ctkt} illustrates our method.
Given $T$ target domains, training is done in $T$ sequential steps.
To ease explanation, we decompose the continual model $F_{(1:t)}$ trained at step $t$ into the feature extractor $F^{\text{feat}}_{(1:t)}$ and the pixel-wise classifier $F^{\text{cls}}_{(1:t)}$.
We denote $F_{(1:t)} =[F^{\text{feat}}_{(1:t)}, F^{\text{cls}}_{(1:t)}]$ with corresponding parameters $\theta_{(1:t)} = [\theta^{\text{feat}}_{(1:t)}, \theta^{\text{cls}}_{(1:t)}]$.

\input{figures/md_ctkt.tex}

\subsection{First step $t=1$}
\label{sec:t1}
In the first iteration, the segmentation model $F_{(1:1)} = F_1 = [F^{\text{feat}}_{(1:1)}, F^{\text{cls}}_{1}]$\footnote{See Sec.\,\ref{sec:t2} for further explanations of this modified notation.} is trained with the traditional UDA approach AdvEnt \cite{vu-cvpr2019} on the single-target problem with the source domain $\mfrk{D}_s$ and the target domain $\mfrk{D}_1$. A discriminator $D_1$ is trained on the source dataset $\domain_s$ and target dataset $\domain_{1}$ by minimizing the classification loss (taking $t=1$):
\begin{equation}
\begin{split}
    \loss_{D_t}& = \frac{1}{|\domain_s|}\sum\limits_{\vx_s\in\domain_s} \loss_\text{BCE}(D_t(\mI_{\vx_s}),1)\\ 
     & \quad + \frac{1}{|\domain_{t}|}\sum\limits_{\vx_{t}\in\domain_{t}} \loss_\text{BCE}(D_t(\mI_{\vx_{t}}),0),\\
\end{split}
\label{eq:loss_d_t}
\end{equation}
where $\loss_\text{BCE}$ is the Binary Cross-Entropy loss and $\mI_{\vx}$ is the weighted self-information map derived from the output $\mP_{\vx}$ of the model as:
\begin{equation}
    \mI_{\vx} = - \mP_{\vx}\,\text{log}\mP_{\vx},
\end{equation}
with pixelwise multiplication and logarithm.
Concurrently, the semantic segmentation model 
%$F_1 = (F^{\text{feat}}_{(1:1)},F^{\text{cls}}_1)$ 
$F_1 = [F^{\text{feat}}_{(1:1)}, F^{\text{cls}}_{1}]$
is trained over its parameters 
%$\theta_1 = (\theta^{\text{feat}}_{(1:1)}, \theta^{\text{cls}}_1)$
$\theta_1=[\theta^{\text{feat}}_{(1:1)}, \theta^{\text{cls}}_1]$
not only to minimize the supervised segmentation loss $\loss_{F_1,\text{seg}}$ on source-domain data, but also to fool the discriminator $D_1$ via minimizing an adversarial loss $\loss_{F_1,\text{adv}}$.
The final objective reads (taking $t=1$):
\begin{equation}
\begin{split}
    \loss_{F_t} &= 
    \underbrace{\frac{1}{|\domain_s|}\sum\limits_{\vx_s\in\domain_s} \loss_\text{seg}(\mP_{\vx_s},\vy_s) }_{\loss_{F_t,\text{seg}}}\\
    &\quad + \lambda_\text{adv}
    \underbrace{\frac{1}{|\domain_{t}|}\sum\limits_{\vx_{t}\in\domain_{t}} \loss_\text{BCE}(D_t(\mI_{\vx_{t}}),1)}_{\loss_{F_t,\text{adv}}},
\end{split}
\label{eq:loss_f_t}
\end{equation}
with a weight $\lambda_\text{adv}$ balancing the two terms; $\loss_\text{seg}$ is the pixel-wise cross-entropy loss.
During training, one alternately minimizes the two losses $\loss_{D_1}$ and $\loss_{F_1}$.

\subsection{Subsequent steps $t\geqslant 2$}
\label{sec:t2}
At iteration $t\geqslant 2$, the segmentation model $F_{(1:t)}$
%, composed of a feature extractor $F^{\mathrm{feat}}_{(1:t)}$ and a pixel-wise classifier $F^{\mathrm{cls}}_{(1:t)}$, 
is trained on the source domain $\mfrk{D}_s$ and the target domain $\mfrk{D}_{t}$ to run on the target domains $1$ to $t$.
%In these subsequent steps, the classification part of the network is designed with multiple branches.
We here re-design the classification part of $F_{(1:t)}$ with two pixel-wise classifiers, referred to as \emph{target-specialist} $F^{\text{cls}}_{t}$ and \emph{target-generalist} $F^{\text{cls}}_{(1:t)}$.
At test time, only the target-generalist $F^{\text{cls}}_{(1:t)}$ is used for prediction. Note that in the special case $t=1$ the target-generalist $F^{\text{cls}}_{(1:1)}$ is equivalent to the target-specialist $F^{\text{cls}}_{1}$ due to dealing with a single target domain. 
%Effectively one can regard the special case of $t=1$ is when the two classifiers are merged as one.
%
\autoref{fig:md_ctkt} provides an overview of the training scheme of $F_{(1:t)}$ when $t\geqslant 2$.

\paragraph{Target-specialist head.}
In this step $t$, the network has a \emph{target $t$-specialist} instrumental segmentation head $F^{\mathrm{cls}}_t$ based on the feature extractor $F^{\mathrm{feat}}_{(1:t)}$. This classifier  handles the specific domain shift between source and target $t$ by performing output-space adversarial alignment between the two domains. % source domain and the target domain $t$. 
This classification head is associated with a domain discriminator $D_t$ to classify source vs. target $t$. The training objectives are similar to those used in single-target models as described for the first iteration (Sec.\,\ref{sec:t1}).
To train the discriminator $D_t$ on the source dataset $\domain_s$ and target dataset $\domain_{t}$, one minimizes the classification loss as defined in \autoref{eq:loss_d_t}.

Concurrently, the target $t$-specialist semantic segmentation model $F_t = [F^{\text{feat}}_{(1:t)},F^{\text{cls}}_t]$ is trained over its parameters $\theta_t = [\theta^{\text{feat}}_{(1:t)}, \theta^{\text{cls}}_t]$ not only to minimize the supervised segmentation loss $\loss_{F_t,\text{seg}}$ on source-domain data, but also to fool the discriminator $D_t$ via minimizing an adversarial loss $\loss_{F_t,\text{adv}}$. The final objective $\loss_{F_t}$ reads as in \autoref{eq:loss_f_t}.
During training, one alternately minimizes the two losses $\loss_{D_t}$ and $\loss_{F_t}$.

\paragraph{Target-generalist head.} The \emph{target-generalist} segmentation head $F^{\mathrm{cls}}_{(1:t)}$, which is eventually kept as classification head of the model, is trained to perform well on all the target domains from $1$ to $t$. The knowledge from the target $t$-specialist branch is distilled to the target-generalist branch via a teacher-student strategy by minimizing the Kullback-Leibler (KL) divergence between the predictions of the two segmentation heads on the target domain $t$. For a given sample $\vx_{t}\in{\domain_{t}}$, we compute the KL loss
\begin{equation}
\resizebox{\linewidth}{!}{%
    $\loss_{\text{KL},t}(\vx_{t}) = \sum\limits_{h = 1}^H\sum\limits_{w = 1}^W\sum\limits_{c = 1}^C\mP_{t,\vx_{t}}{\scriptstyle [h,w,c]} \log \frac{\mP_{t,\vx_{t}}{\scriptstyle [h,w,c]}}{\mP_{(1:t),\vx_{t}}{\scriptstyle [h,w,c]}},$
}
\end{equation}
where $H\,{\times}\,W$ is image dimension;  $\mP_{t,\vx_{t}}$ and $\mP_{(1:t),\vx_{t}}$ are soft-segmentation predictions coming from the target-specialist $F^{\mathrm{cls}}_t$ and the target-generalist $F^{\mathrm{cls}}_{(1:t)}$ respectively. 

\subsection{Preventing catastrophic forgetting} 

Furthermore, the model must not forget what it has learned in the previous training iterations about the other target domains from $1$ to $t-1$. Without proper constraint, the model may be subject to catastrophic forgetting, degrading its performance on previous target domains. Thus, we consider additional losses based on the model's previous iteration when training the model on the new target domain. The aim of these losses is to make sure that the new model keeps similar features to its previous iteration in order to keep similar results on the previous target domains, on which the old model was performing well.

Ideally, one would want to use images from the previous target domains to get the features and results of the old model and make sure they are as close as possible to those of the currently-trained model. However, in our continual setting, one cannot access images of the previous domains anymore.  %Nevertheless, since the models are trained with
Thanks to adversarial alignment, we here assume that the features on the source domain are close enough to the features of the previous target domains to use them as proxy. Under this assumption, we constrain source domain features of the current model with source domain features of the previous model. 

Practically, we perform a similar distribution distillation with KL distillation between the previously-trained 
%target-generalist head $F^{\mathrm{cls}}_{(1:t-1)}$, 
segmentation model $F_{(1:t-1)}$, 
supposed to perform well on targets $1$ to $t-1$, and the currently-trained 
%target-generalist head $F^{\mathrm{cls}}_{(1:t)}$
segmentation model $F_{(1:t)} = [F^{\mathrm{feat}}_{(1:t)},F^{\mathrm{cls}}_{(1:t)}]$ 
we want to handle all targets $1$ to $t$. Without access to images from domains $1$ to $t-1$, we use source images as proxy. For a given source sample $\vx_s\in{\domain_s}$, we compute the KL loss
\begin{equation}
\resizebox{\linewidth}{!}{%
    $\loss_{\text{KL},(1:t-1)}(\vx_s) = \sum\limits_{h = 1}^H\sum\limits_{w = 1}^W\sum\limits_{c = 1}^C\mP_{(1:t-1),\vx_s}{\scriptstyle [h,w,c]} \log \frac{\mP_{(1:t-1),\vx_s}{\scriptstyle [h,w,c]}}{\mP_{(1:t),\vx_s}{\scriptstyle [h,w,c]}},$
}
\end{equation}
where $\mP_{(1:t-1),\vx_s}$ and $\mP_{(1:t),\vx_s}$ are soft-segmentation predictions coming from the previous iteration of the target-generalist classifier $F^{\mathrm{cls}}_{(1:t-1)}$, previously trained for targets $1$ to $t-1$ and now frozen, and the currently-trained target-generalist head $F^{\mathrm{cls}}_{(1:t)}$, respectively.

With the addition of this loss, the minimization objective of the target-generalist model $F_{(1:t)} = [F^{\mathrm{feat}}_{(1:t)},F^{\mathrm{cls}}_{(1:t)}]$ over its parameters $\theta_{(1:t)} = [\theta^{\mathrm{feat}}_{(1:t)},\theta^{\mathrm{cls}}_{(1:t)}]$, or distribution distillation loss, then reads:
\begin{equation}
\begin{split}
    \loss_{F_{(1:t)}} &= 
    \frac{1}{|\domain_{t}|}
    \sum\limits_{\vx_{t}\in\domain_{t}}
    \loss_{\text{KL},t}(\vx_{t})\\
    &\quad + \lambda_{\text{prev}} \frac{1}{|\domain_s|} \sum\limits_{\vx_s\in\domain_s}
    \loss_{\text{KL},(1:t-1)}(\vx_s),
\end{split}
\end{equation}
with weight $\lambda_{\text{prev}}$ to balance the distribution distillation from the previous model.

Along with this distribution distillation loss, we also want to specifically enforce the current feature extractor $F^{\mathrm{feat}}$ to produce features at each layer close to the previous iteration of the feature extractor. This kind of approach helps alleviate the catastrophic forgetting in continual learning problems. Practically, we propose to use the Local POD distillation $\loss_\text{LocalPod}(\theta^{\mathrm{feat}})$ proposed in \cite{douillard2021plop}. Denoting $\psi(\vu)$ the concatenation of width-pooled slices and height-pooled slices over multiple scales of $\vu$, the Local POD distillation loss is defined, on each sample $\vx_s$ of the source dataset $\domain_s$, and over each layer activation $F^{\mathrm{feat\,}(l)}_{(1:t)}(\vx_s)$ and $F^{\mathrm{feat\,}(l)}_{(1:t-1)}(\vx_s)$, $l\in[L]$, of the the feature extractors $F^{\mathrm{feat}}_{(1:t)}$ and $F^{\mathrm{feat}}_{(1:t-1)}$, as:
\begin{equation}
    \loss_{\text{LocalPod}}(\theta^{\text{feat}}) = \frac{1}{L}\sum\limits_{l=1}^L \left\lVert \psi(F^{\mathrm{feat\,}(l)}_{(1:t)}) - \psi(F^{\mathrm{feat\,}(l)}_{(1:t-1)}) \right\rVert^2_F.
\end{equation}
It is a multi-scale pooling distillation method that aims to preserve spatial relationships on the intermediate features. A more in-depth definition of the loss is developed in \cite{douillard2021plop}.

Note that this distillation loss was originally proposed in \cite{douillard2021plop} in a continual semantic segmentation setting where new classes were added at each iteration while staying in the same domain. The setting considered here is notably different since the classes do not change throughout training, but the model encounters new target domains at each iteration. Furthermore, the target domains are not labeled, while segmentation maps are available for all images during training in the setting of \cite{douillard2021plop}. 

Overall, the minimization objective of the semantic segmentation model $F$ over $\theta$ can be written as:
\begin{equation}
    \loss_F = \loss_{F_t,\text{seg}} + \lambda_{\text{adv}} \loss_{F_t,\text{adv}} + \lambda_{\text{dd}} \loss_{F_{(1:t)}} + \lambda_{\text{fd}} \loss_{\text{LocalPod}},
    \label{eq:ctkt_l_f}
\end{equation}
with factors $\lambda_{\text{adv}}, \lambda_{\text{dd}}, \lambda_{\text{fd}}$ balancing adversarial training, distribution distillation and feature distillation, respectively.

\section{Experimental Results}
\subsection{Experimental Setting}
\paragraph{Continual Protocol.}
Each continual UDA experiment follows the same training protocol: when training a model on a continual benchmark $\mfrk{D}_s\,\shortrightarrow\mfrk{D}_1\,\shortrightarrow\mfrk{D}_2\,\shortrightarrow\dots\,\shortrightarrow\mfrk{D}_T$, the model trains in $T$ steps. At each step $t$, the model is initialized from the previous model at iteration $t-1$ and is trained in a UDA fashion on $\mfrk{D}_s\,\shortrightarrow\mfrk{D}_t$, i.e. on the labeled source domain $\mfrk{D}_s$ and the unlabeled target domain $\mfrk{D}_t$. Note that the model leverages at each step the supervision from the same source domain and that it cannot train again on the target domains it has previously seen.
\paragraph{Datasets.}
We build our experiments on four popular semantic segmentation datasets.
GTA5 \cite{richter-eecv2016} is a dataset of 24,966 synthetic images of size $1920\times 1080$ rendered using the eponymous open-world video game. 
    %The urban scenes are all from the car perspective and mimic the streets of American cities. Every image is labeled with a pixel-wise semantic segmentation map with labels over 19 classes. This dataset is commonly used as a source domain for UDA experiments. In the continual unsupervised domain adaptation context, we
    We 
    consider GTA5 as source domain and its training dataset is always available.
Cityscapes \cite{cordts-cvpr2016} contains $2048 \times 1024$-sized labeled urban scene images from cities around Germany, split in training and validation sets of 2,975 and 500 samples respectively. 
    %Human annotations are based on 34 classes, divided in 7 high-level categories (or super classes). While the segmentation maps for training examples are available, note that they are not used in UDA applications where Cityscapes is used as a target domain.
Mapillary Vistas \cite{neuhold-iccv2017}, Mapillary in short, is a dataset collected in multiple cities around the world, which is composed of 18,000 training and 2,000 validation labeled scenes of varying sizes.
IDD \cite{varma-wacv2019} is an Indian urban dataset having 6,993 training and 981 validation labeled scenes of size $1280\times 720$.

As in \cite{saporta2021mtaf}, we standardize the label set with the 7 super classes common to all four datasets. In this continual unsupervised domain adaptation settings, we train the model sequentially on the target datasets, one after the other. Nonetheless, evaluation of continual models is performed on all the seen target datasets.

\paragraph{Implementation Details.}
The experiments are conducted with PyTorch \cite{paszke-nips2017}.
The adversarial framework is based on AdvEnt's published code\footnote{\url{https://github.com/valeoai/ADVENT}} \cite{vu-cvpr2019}.
The semantic segmentation model is DeepLab-V2 \cite{chen-tpami2018}, built upon the ResNet-101 \cite{he-cvpr2016} backbone first initialized with ImageNet \cite{deng-cvpr2009} pre-trained weights. In a continual setting, when considering a new target domain, the model is initialized with the weights of the previously-trained model.
All semantic segmentation models are trained by SGD \cite{bottou2010large} with learning rate $2.5\cdot 10^{-4}$, momentum $0.9$ and weight decay $10^{-4}$.
We train the discriminators using an Adam optimizer \cite{kingma-iclr2015} with learning rate $10^{-4}$. All experiments are conducted at the $640 \times 320$ resolution.
For MuHDi, the weights $\lambda_{\text{prev}}$ and $\lambda_{\text{dist}}$, balancing the distribution distillation from the previously-trained model, are set to $10^{-5}$ in all experiments. We use the authors' implementation of the Local POD loss~\footnote{\url{https://github.com/arthurdouillard/CVPR2021_PLOP}}.

\input{tables/ctkt_G-CI.tex}

\paragraph{Evaluation.}
Semantic segmentation models are generally evaluated in terms of Intersection over Union (IoU), also known as Jaccard index, per class or mean Intersection over Union (mIoU) over all classes, expressed in percentage. In our specific scenario where models are evaluated on multiple target domains, we also include the mIoU averaged over the target domains (mIoU Avg.) as in \cite{saporta2021mtaf}.

\input{tables/ctkt_ablation.tex}

\input{tables/ctkt_G-CIMv2.tex}

\subsection{Two-Target Domain Results on GTA5 $\shortrightarrow$ Cityscapes $\shortrightarrow$ IDD}
We first consider the two-step continual task GTA5 $\shortrightarrow$ Cityscapes $\shortrightarrow$ IDD: the models are first trained on GTA5 $\shortrightarrow$ Cityscapes, then on GTA5 $\shortrightarrow$ IDD. \autoref{tab:ctkt_results_2t} shows the results on this benchmark. 
First, the results of the continual baseline highlight the catastrophic forgetting problem. Indeed, while the model on the previous step (GTA5 $\shortrightarrow$ Cityscapes) has a $69.0\%$ mIoU score on Cityscapes, the performance on Cityscapes of the continual baseline drops to $63.6\%$ mIoU ($-5.4\%$). MuHDi, which directly tackles catastrophic forgetting in its design, only loses $1.0\%$ mIoU on Cityscapes from the previous step while having similar performance on the new target domain IDD compared to the continual baseline. Moreover, the performance of MuHDi in terms of mIoU Avg. gets close to the multi-target models, which serve as oracles, further demonstrating its efficiency.

\paragraph{Ablation Study.} We perform an ablation study on this setting of the proposed method MuHDi after this second training step. 
The results are displayed in \autoref{tab:ctkt_ablation}.

In this table, we analyze the impact of the multiple elements of MuHDi: probability distribution distillation from the previous model (`Distribution Distillation'), feature distillation from the previous model with $\loss_\text{LocalPod}$ (`Feature Distillation'), and the decomposition into target-specialist and target-generalist branches (`Multi Heads').

First, we note that all the continual experiments exhibit comparable performance on IDD, on which they were trained last: from $65.2\%$ to $65.5\%$ mIoU on IDD, much higher than the $61.5\%$ mIoU on IDD of the previous model, not trained on this domain.

Then, we note that the continual baseline (second row of \autoref{tab:ctkt_ablation} with $65.5\%$ mIoU average), not implementing any of these elements, suffers catastrophic forgetting on the previous Cityscapes target domain, dropping from $69.0\%$ to $63.6\%$ mIoU. Despite the improvement on IDD, on which it was trained last, the overall performance is lower than the previous model (difference of $-0.7\%$ mIoU Avg.).

Implemented on their own, both the distribution distillation from the previous model and the feature distillation with LocalPod help limit the catastrophic forgetting on Cityscapes. Furthermore, they prove to be complementary, exhibiting better results on Cityscapes when combined.

Finally, MuHDi adds the decomposition into target-specific and target-agnostic branches, which lessens catastrophic forgetting even further. MuHDi performance on Cityscapes is only $1.0\%$ mIoU lower than the previous model trained only on Cityscapes while being competitive on IDD. Overall, the performance of MuHDi in terms of mIoU Avg. is significantly higher than the other models, notably $+2.2\%$ greater than the continual baseline.

\subsection{Three-Target Domain Results on GTA5 $\shortrightarrow$ Cityscapes $\shortrightarrow$ IDD $\shortrightarrow$ Mapillary}
We consider the challenging setup involving three target domains -- Cityscapes, Mapillary and IDD -- discovered sequentially in a three-step continual learning setting and show results in \autoref{tab:ctkt_results}.  We compare results of continual UDA with multi-target UDA from \cite{saporta2021mtaf}. Due to the simultaneous availability of data for all the target domains, the multi-target setting is easier than the continual setting and the performance of multi-target experiments are expected to be higher than those of continual UDA experiments.

The continual baseline performs worse than all the multi-target models with at least a $-0.8\%$ mIoU Avg. decrease. In particular, the results on the Cityscapes and IDD datasets, which have been seen in early continual training steps, are significantly lower than those of all the other models due to catastrophic forgetting. Moreover, its performance is notably degraded on the \emph{human} and \emph{vehicle} classes compared to the better performing multi-target models, which is especially critical for autonomous driving applications.

On the other hand, MuHDi exhibits comparable performance to the rather competitive Multi-Dis.\,multi-target model \cite{saporta2021mtaf} with a $68.2\%$ mIoU Avg., proving the efficiency of the proposed continual learning framework.

\section{Conclusion}
Practical applications of UDA require models that perform on a multitude of different domains, such as multiple cities or various weather conditions. While effective in the traditional single-target setting, standard UDA strategies do not easily improve and learn new target domains when confronted to new environments.

We proposed MuHDi, for Multi-Head Distillation, to tackle the continual UDA problem. MuHDi adapts to each new target domain using a probability distribution distillation strategy from a target-specialist head to a target-generalist head, while also distilling the distribution and features from the previous model, trained to perform on all the previous target domains. 
The proposed benchmarks and architecture deliver competitive baselines for future developments of real-world %use-cases
UDA scenarios like continual UDA.

\vspace*{0.cm}{\small\noindent\textbf{Acknowledgments}: This work was supported by ANR grant VISA DEEP (ANR-20-CHIA-0022), and HPC resources of IDRIS under the allocation 2021-[AD011012153R1] made by GENCI.}

%%%%%%%%% REFERENCES
{\small
\bibliographystyle{ieee_fullname}
\bibliography{egbib}
}

\end{document}

%% file: macro.tex
\definecolor{ggreen}{RGB}{15,157,88}
\definecolor{gred}{RGB}{219,68,55}
\newcolumntype{R}[2]{%
    >{\adjustbox{angle=#1,lap=\width-(#2)}\bgroup}%
    l%
    <{\egroup}%
}
\newcommand*\rot{\multicolumn{1}{R{45}{1em}}}

\newcommand{\mcl}[1]{\ensuremath{\mathcal{#1}}}

\newcommand{\mbb}[1]{\ensuremath{\mathbb{#1}}}

\newcommand{\mfrk}[1]{\ensuremath{\mathfrak{#1}}}
\newcommand{\ve}[1]{\ensuremath{\bm{#1}}} % vectors
\newcommand{\m}[1]{\ensuremath{\bm{#1}}} % matrices
\newcommand{\s}[1]{\ensuremath{\mathcal{#1}}} % sets

\def\mI{\m{I}}

\def\mP{\m{P}}

\def\vu{\ve{u}}
\def\vx{\ve{x}}
\def\vy{\ve{y}}

\def\loss{\mcl{L}}
\def\domain{\s{X}}

\definecolor{flat}{RGB}{128,64,128}
\definecolor{construction}{RGB}{70,70,70}
\definecolor{object}{RGB}{220,220,0}
\definecolor{nature}{RGB}{107,142,35}
\definecolor{sky}{RGB}{70,130,180}
\definecolor{human}{RGB}{220,20,60}
\definecolor{vehicle}{RGB}{0,0,142}

%% file: figures/bigpicture.tex
\begin{figure}[t!]
    \centering
     \includegraphics[trim=4mm 0 3mm 0,clip,width=\linewidth]{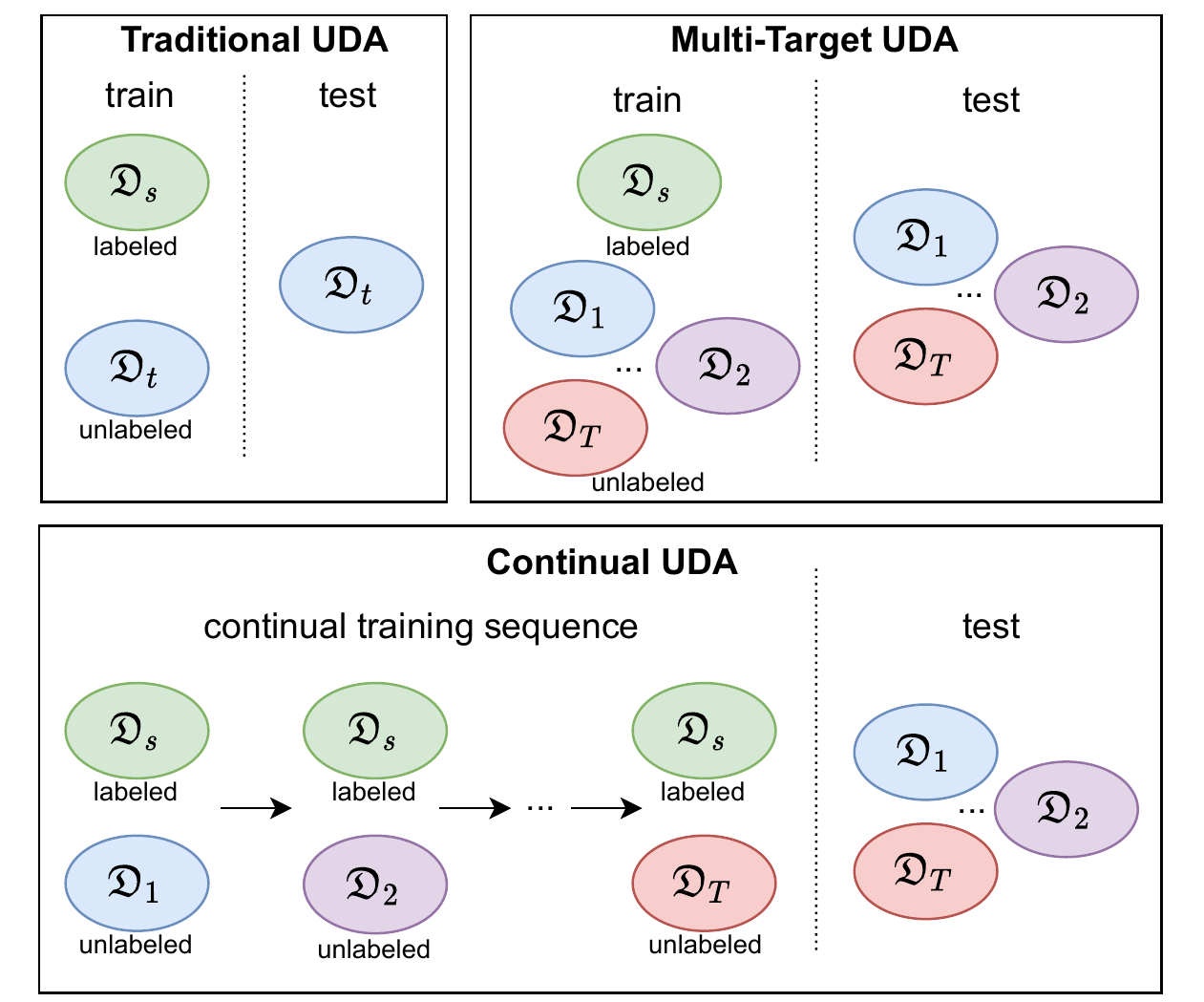}
    \caption{Traditional UDA trains a model on an unlabeled target domain $\mfrk{D}_t$ leveraging a labeled source domain $\mfrk{D}_s$. Multi-target UDA trains a model on multiple unlabeled target domains $\mfrk{D}_1$, $\mfrk{D}_2$,... $\mfrk{D}_T$ simultaneously leveraging a labeled source domain $\mfrk{D}_s$. Continual UDA is a more realistic and challenging setting in which the model sequentially learns unlabeled target domains $\mfrk{D}_1$, $\mfrk{D}_2$,... $\mfrk{D}_T$, without access of the previously seen target domains, by leveraging a labeled source domain $\mfrk{D}_s$.}
    \label{fig:bigpicture}
\end{figure}

%% file: figures/md_ctkt.tex
\begin{figure*}[t!]
    \centering
     \includegraphics[trim=0 0 12mm 0,clip,width=.85\textwidth]{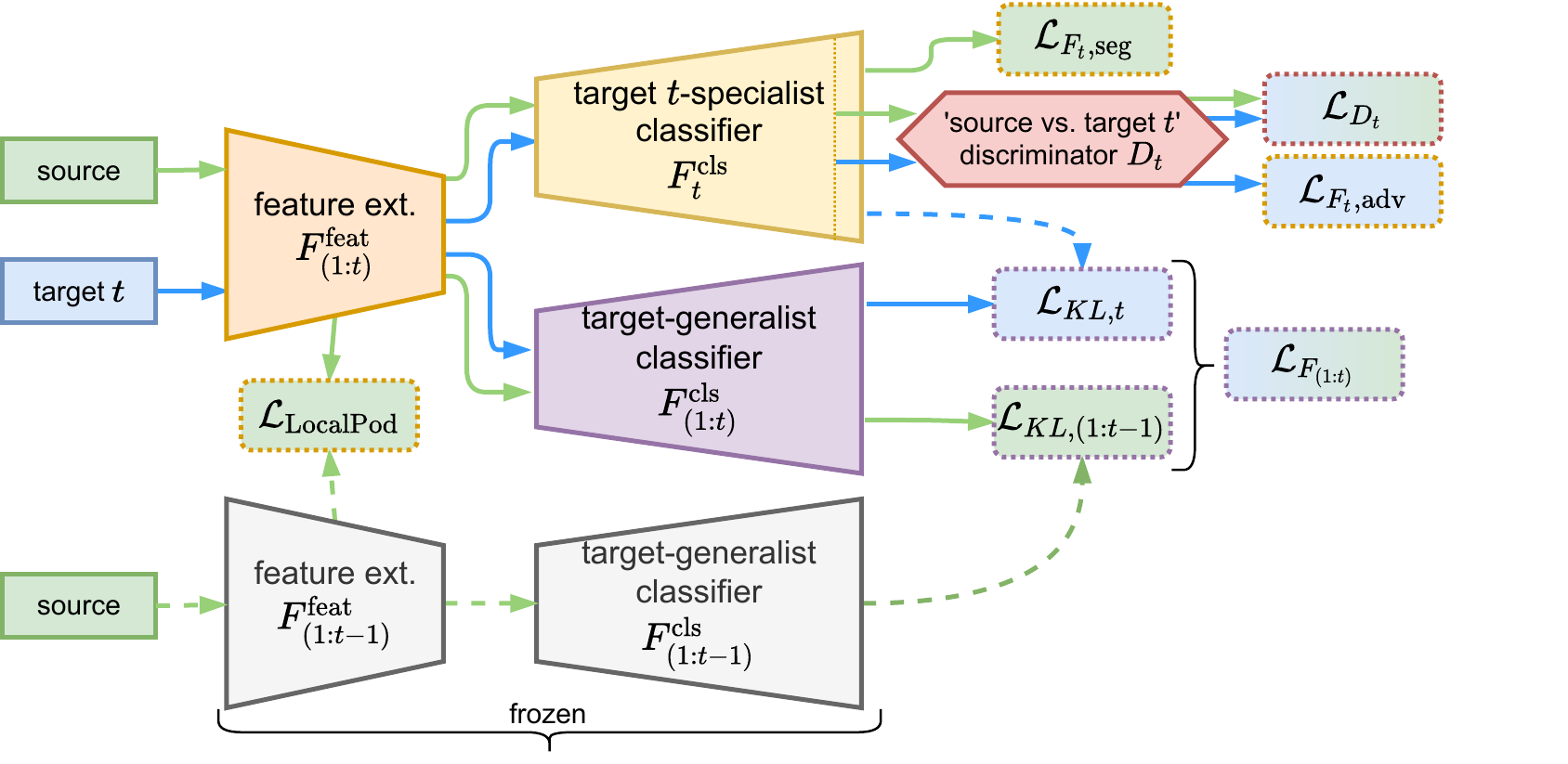}
    \caption{\textbf{MuHDi approach to continual UDA.} 
    When discovering a new target domain $t$, MuHDi learns its target-generalist model $F_{(1:t)}=[F^{\text{feat}}_{(1:t)}, F^{\text{cls}}_{(1:t)}]$ using knowledge distillation from both a target $t$-specialist segmentation head $F^{\text{cls}}_t$ trained adversarially for this target domain, and the frozen segmentation model from the previous training step $F_{(1:t-1)}=[F^{\text{feat}}_{(1:t-1)}, F^{\text{cls}}_{(1:t-1)}]$. In combination with this architectural design, the training losses are indicated and further developed in the text. In particular, the Local POD loss $\loss_{\text{LocalPod}}$ is introduced to further prevent catastrophic forgetting. The losses are not back-propagated into dotted arrows.}
    \vspace{-0.2cm}
    \label{fig:md_ctkt}
\end{figure*}

%% file: tables/ctkt_G-CI.tex
\begin{table}[t!]
    \centering
    \resizebox{.9\linewidth}{!}{%
            \begin{tabular}{c| l | l| l|l<{\kern-\tabcolsep}}
                \toprule
                \multicolumn{5}{c}{\textbf{GTA5\,$\shortrightarrow$\,Cityscapes\,$\shortrightarrow$\,IDD}}\\
                \midrule
                Setting & Method & Target & mIoU&\shortstack[l]{mIoU\\Avg.}\\
                \midrule
                \multirow{6}{*}{\rotatebox[origin=c]{90}{\shortstack[c]{Multi-Target\\Oracle}}}&\multirow{2}{*}{\shortstack[l]{Multi-Target\\Baseline \cite{vu-cvpr2019}}} & Cityscapes & 70.0& \multirow{2}{*}{67.4}  \\
                & & IDD &	64.8 & \\
                \cmidrule{2-5}
                &\multirow{2}{*}{Multi-Dis. \cite{saporta2021mtaf}}  & Cityscapes & 68.9 & \multirow{3}{*}{67.3}\\
                & & IDD & 	65.7 &  \\
                \cmidrule{2-5}
                &\multirow{2}{*}{MTKT \cite{saporta2021mtaf}} &Cityscapes &  70.4 & \multirow{2}{*}{68.2} \\ 
                & & IDD & 65.9 &   \\
                \midrule
                \multirow{4}{*}{\rotatebox[origin=c]{90}{Continual}}&\multirow{2}{*}{\shortstack[l]{Continual\\Baseline \cite{vu-cvpr2019}}}  & Cityscapes & 63.6  & \multirow{2}{*}{64.5}  \\
                & & IDD & \textbf{65.4} & \\
                \cmidrule{2-5}
                &\multirow{2}{*}{MuHDi (ours)}  & Cityscapes & \textbf{68.0} & \multirow{2}{*}{\textbf{66.7}}  \\
                & & IDD &  \textbf{65.3} & \\
                \bottomrule
            \end{tabular}
        }
    \caption{\textbf{Continual UDA segmentation performance on GTA5 $\shortrightarrow$ Cityscapes $\shortrightarrow$ IDD (two steps).} `Setting' indicates if the experiment is multi-target (oracle, simultaneous training on all the target domains) or continual (target domains discovered sequentially, two-step training). \textbf{Bold} indicates the best continual performance in terms of mIoU.
    }
    \vspace{-0.2cm}
    \label{tab:ctkt_results_2t}
    \end{table}

%% file: tables/ctkt_ablation.tex
\begin{table}[t!]
    \centering
	\resizebox{.9\linewidth}{!}{%
        \begin{tabular}{l|ccc|ccc}
            %\toprule
            %\multicolumn{7}{c}{\textbf{GTA5\,$\shortrightarrow$\,Cityscapes\,$\shortrightarrow$\,IDD}}\\
            %\midrule
            %\midrule
            % & \multicolumn{3}{c}{Method} & \multicolumn{3}{c}{Eval on} \\
            %\cmidrule{1-2} \cmidrule{4-5}  \cmidrule{7-9}\\
            \multicolumn{1}{c}{Training Step} %Training Step 
            & \rot{\footnotesize{\textbf{Distribution Distillation}}} & \rot{\footnotesize{\textbf{Feature Distillation}}} & \rot{\footnotesize{\textbf{Multi Heads}}} & \shortstack[c]{mIoU\\ Cityscapes} & \shortstack[c]{mIoU\\ IDD} & \shortstack[c]{mIoU\\ Avg.} \\
            \midrule
            $t=1$ \quad G\,$\shortrightarrow\,$C & & & & \textbf{69.0} & 61.5 & 65.2 \\
            \midrule
            \multirow{5}{*}{$t=2$ \quad G\,$\shortrightarrow\,$C\,$\shortrightarrow\,$I} & & & & 63.6 & \underline{65.4} & 64.5 \\
             & \checkmark & & & 65.3 & 65.3 & 65.3 \\
             & & \checkmark & & 66.4 & \textbf{65.5} & 65.9 \\
             & \checkmark & \checkmark & & 66.8 & \textbf{65.5} & \underline{66.2} \\
             & \checkmark & \checkmark & \checkmark & \underline{68.0} & 65.3 & \textbf{66.7} \\
            \bottomrule
        \end{tabular}%
    }
\caption{\textbf{Ablation study of the MuHDi architecture for continual UDA.} All the models are trained in a similar fashion in the first training step ($t=1$) on GTA5\,$\shortrightarrow$\,Cityscapes (`G\,$\shortrightarrow\,$C') with AdvEnt \cite{vu-cvpr2019}. In the second step ($t=2$), the models are trained on GTA5\,$\shortrightarrow$\,IDD from the previous iteration (`G\,$\shortrightarrow\,$C\,$\shortrightarrow\,$I') and are decomposed into multiple blocks
%: `Distribution Distillation' denotes distribution distillation from the previous model; `Feature Distillation' denotes feature distillation with Local POD \cite{douillard2021plop}; `Multi Heads' denotes training both a target-specialist head with adversarial training and a target-generalist head with knowledge distillation. 
. 
%The best results for each metric are in \textbf{bold}, the second best are \underline{underlined}.
}
\vspace{-0.2cm}
\label{tab:ctkt_ablation}    
\end{table}

%% file: tables/ctkt_G-CIMv2.tex
\begin{table*}[ht!]
    \centering
    \resizebox{.9\textwidth}{!}{%
            \begin{tabular}{c| l | l| c c c c c c c|l|l<{\kern-\tabcolsep}}
                \toprule
                \multicolumn{12}{c}{\textbf{GTA5\,$\shortrightarrow$\,Cityscapes\,$\shortrightarrow$\,IDD\,$\shortrightarrow$\,Mapillary}}\\
                \midrule
                Setting & Method & Target  & \rotatebox{90}{flat} & \rotatebox{90}{constr.} & \rotatebox{90}{object} & \rotatebox{90}{nature} & \rotatebox{90}{sky} & \rotatebox{90}{human} & \rotatebox{90}{vehicle\,} & mIoU&\shortstack[l]{mIoU\\Avg.}\\
                \midrule
                \multirow{9}{*}{\rotatebox[origin=c]{90}{\shortstack[c]{Multi-Target\\Oracle}}}&\multirow{3}{*}{\shortstack[l]{Multi-Target\\Baseline \cite{vu-cvpr2019}}} & Cityscapes &  93.6 &	80.6 &	26.4 &	78.1 &	81.5 &	51.9 &	76.4 &	69.8 & \multirow{3}{*}{67.8}  \\
                & & IDD &  92.0 &	54.6 &	15.7 &	77.2 &	90.5 &	50.8 &	78.6 &	65.6 & \\
                & & Mapillary &  89.2 &	72.4 &	32.4 &	73.0 &	92.7 &	41.6 &	74.9 &	68.0 & \\
                \cmidrule{2-12}
                &\multirow{3}{*}{Multi-Dis. \cite{saporta2021mtaf}}  & Cityscapes &  94.6 &	80.0 &	20.6 &	79.3 &	84.1 &	44.6 &	78.2 &	68.8 & \multirow{3}{*}{68.2}\\
                & & IDD &  91.6 &	54.2 &	13.1 &	78.4 &	93.1 &	49.6 &	80.3 &	65.8 &  \\
                & & Mapillary &  89.0 &	72.5 &	29.3 &	75.5 &	94.7 &	50.3 &	78.9 &	70.0 & \\
                \cmidrule{2-12}
                &\multirow{3}{*}{MTKT \cite{saporta2021mtaf}} &Cityscapes &  94.6 &	80.7 &	23.8 &	79.0 &	84.5 &	51.0 &	79.2 &	70.4 & \multirow{3}{*}{69.1} \\ 
                & & IDD &  91.7 &	55.6 &	14.5 &	78.0 &	92.6 &	49.8 &	79.4 &	65.9 &   \\
                & & Mapillary &  90.5 &	73.7 &	32.5 &	75.5 &	94.3 &	51.2 &	80.2 &	71.1 & \\
                
                \midrule
                \multirow{6}{*}{\rotatebox[origin=c]{90}{Continual}}&\multirow{3}{*}{\shortstack[l]{Continual\\Baseline \cite{vu-cvpr2019}}}  & Cityscapes & 92.9 & 79.0 & 18.7 & 76.9 & 84.1 & 47.3 & 72.9 & 67.4  & \multirow{3}{*}{67.0}  \\
                & & IDD & 91.8 & 51.1 & 11.6 &79.0 &91.6 &47.5 &72.5 &63.6 & \\
                & & Mapillary &  90.3 & 71.7 & 30.1 & 76.1 & 93.9 & 50.2 & 77.3 & \textbf{70.0} & \\
                \cmidrule{2-12}
                &  & Cityscapes &  94.9 & 80.2 & 19.3 & 79.4 & 80.7 & 53.2 & 78.2 & \textbf{69.4} &   \\
                & & IDD &  92.5 &54.1 &12.0 & 79.2 & 92.7 & 48.0 & 76.6 & \textbf{65.0} & \\
                &\multirow{-3}{*}{MuHDi (ours)}& Mapillary &  91.0 & 73.2 & 29.2 & 76.0 & 94.1 & 50.0 & 78.0 & \textbf{70.2} & \multirow{-3}{*}{\textbf{68.2}}\\
                \bottomrule
            \end{tabular}
        }
    \vspace{-0.2cm}
    \caption{\textbf{Continual UDA segmentation performance on GTA5 $\shortrightarrow$ Cityscapes $\shortrightarrow$ IDD $\shortrightarrow$ Mapillary (three steps).} `Setting' indicates if the experiment is multi-target (oracle, simultaneous training on all the target domains) or continual (target domains discovered sequentially, three-step training). \textbf{Bold} indicates the best continual performance in terms of mIoU.
    }
    \vspace{-0.2cm}
    \label{tab:ctkt_results}
    \end{table*}